\documentclass[10.5pt,twocolumn,journal,letterpaper]{IEEEtran}
\usepackage{amsmath, amsthm, amssymb}
\usepackage{url,flushend,multirow,booktabs}
\usepackage{algorithm,algorithmic}   
\usepackage{graphicx}
\usepackage{bm}
\usepackage{cite}
\usepackage{subfigure}
\usepackage{float}
\usepackage{color}
\usepackage{multirow}

\usepackage{colortbl}
\definecolor{maroon}{cmyk}{0.08,0.04,0.00,0.06}  
\definecolor{citecol}{RGB}{0,220,0}  

\usepackage[colorlinks,citecolor=citecol,urlcolor=blue,bookmarks=false,hypertexnames=true]{hyperref} 

\newcommand{\eg}{e.g.}
\newcommand{\ie}{i.e.}

\newcommand{\etal}{\textit{et al.}}
\newcommand{\aka}{\textit{a.k.a.}}

\DeclareFixedFont{\mf}{OT1}{ptm}{m}{n}{10pt}
\DeclareFixedFont{\mfb}{OT1}{ptm}{bx}{n}{10pt}

\begin{document}
%

\title{Triple-View Knowledge Distillation for Semi-Supervised Semantic Segmentation}
%
%
%

\author{Ping~Li,~\IEEEmembership{Member,~IEEE}, Junjie~Chen, Li~Yuan, Xianghua~Xu, and Mingli~Song,~\IEEEmembership{Senior Member,~IEEE},
\thanks{P.~Li, J.~Chen and X.~Xu are with the School of Computer Science and Technology, Hangzhou Dianzi University, Hangzhou, China (e-mail: lpcs@hdu.edu.cn, cjj@hdu.edu.cn, xhxu@hdu.edu.cn). P.~Li is also with Guangdong Laboratory of Artificial Intelligence and Digital Economy (SZ), China.}
\thanks{L.~Yuan is with the School of Electronic and Computer Engineering, Peking University, China, and also with PengCheng Laboratory, China (e-mail:yuanli-ece@pku.edu.cn).}
\thanks{M.~Song is with the College of Computer Science, Zhejiang University, Hangzhou, China. (e-mail:songml@zju.edu.cn).} 
}
\markboth{arXiv}
{LI \MakeLowercase{\textit{et al.}}:~Triple-View Knowledge Distillation for Semi-Supervised Semantic Segmentation}
%

\maketitle

\begin{abstract}
  To alleviate the expensive human labeling, semi-supervised semantic segmentation employs a few labeled images and an abundant of unlabeled images to predict the pixel-level label map with the same size. Previous methods often adopt co-training using two convolutional networks with the same architecture but different initialization, which fails to capture the sufficiently diverse features. This motivates us to use tri-training and develop the triple-view encoder to utilize the encoders with different architectures to derive diverse features, and exploit the knowledge distillation skill to learn the complementary semantics among these encoders. Moreover, existing methods simply concatenate the features from both encoder and decoder, resulting in redundant features that require large memory cost. This inspires us to devise a dual-frequency decoder that selects those important features by projecting the features from the spatial domain to the frequency domain, where the dual-frequency channel attention mechanism is introduced to model the feature importance. Therefore, we propose a Triple-view Knowledge Distillation framework, termed TriKD, for semi-supervised semantic segmentation, including the triple-view encoder and the dual-frequency decoder. Extensive experiments were conducted on two benchmarks, \ie, Pascal VOC 2012 and Cityscapes, whose results verify the superiority of the proposed method with a good tradeoff between precision and inference speed. 
 
\end{abstract}


 \ifCLASSOPTIONpeerreview
 \begin{center} \bfseries EDICS Category: 3-BBND \end{center}
 \fi

\IEEEpeerreviewmaketitle

\section{Introduction}
\label{sec1:intro}

\IEEEPARstart{S}{emantic} segmentation predicts the pixel-level label for image, and has many applications, \eg, autonomous driving~\cite{caesar-cvpr2020-nuscenes} and scene understanding~\cite{zhou-ijcv2019-semantic}. While much progress has been achieved by previous fully-supervised methods, it requires large human costs for pixel-level labeling of a huge number of images. This encourages the exploration of semi-supervised semantic segmentation \cite{chen-cvpr2021-cps}, which employs only a few labeled images and a plentiful unlabeled images during model training. Existing methods are roughly divided into two categories, including \textit{self-training} \cite{wang-cvpr2022-u2pl} and \textit{consistency regularization} \cite{chen-cvpr2021-cps}. The former employs the labeled samples to build a teacher network that generates the pseudo-labels of unlabeled samples, which are added to the training set for learning a student network. It progressively updates the initial model by iteration, and the pseudo-labels with possible noise may result in the noisy training samples, thus obtaining an inferior student network. By contrast, the latter emphasizes the output consistency of the model under various perturbations, including image-level \cite{yang-cvpr2022-st++}, feature-level \cite{ouali-cvpr2020-cct}, and network-level \cite{chen-cvpr2021-cps, ke-eccv2020-gct} perturbation. This work belongs to the latter and focuses on the network-level perturbation.

Existing network-level methods usually train two models with the same structure but different initialization, which can be regarded as co-training \cite{qiao-eccv2018-cotraining}. In particular, the output of one model supervises the other model as the pseudo-label, but it cannot guarantee the diverse and complementary outputs from the two models, due to their possible inconsistency during training. Moreover, semantic segmentation desires a large receptive field to capture the global context of image, which is not well satisfied by previous methods \cite{chen-cvpr2021-cps,ke-eccv2020-gct} that only adopt Convolutional Neural Network (CNN) \cite{gu-pr2018-cnn} as the backbone. Worse still, convolutional networks have the inductive bias, \ie, locality and translation equivariance \cite{dosovitskiy-iclr2021-vit,zheng-cvpr2021-setr}, which limits the receptive field of the model and constrains its ability of capturing the global context by reflecting the long-range pixel dependency \cite{zheng-cvpr2021-setr}.

\begin{figure}[!t]
	\centering
	\includegraphics[width=0.9\linewidth]{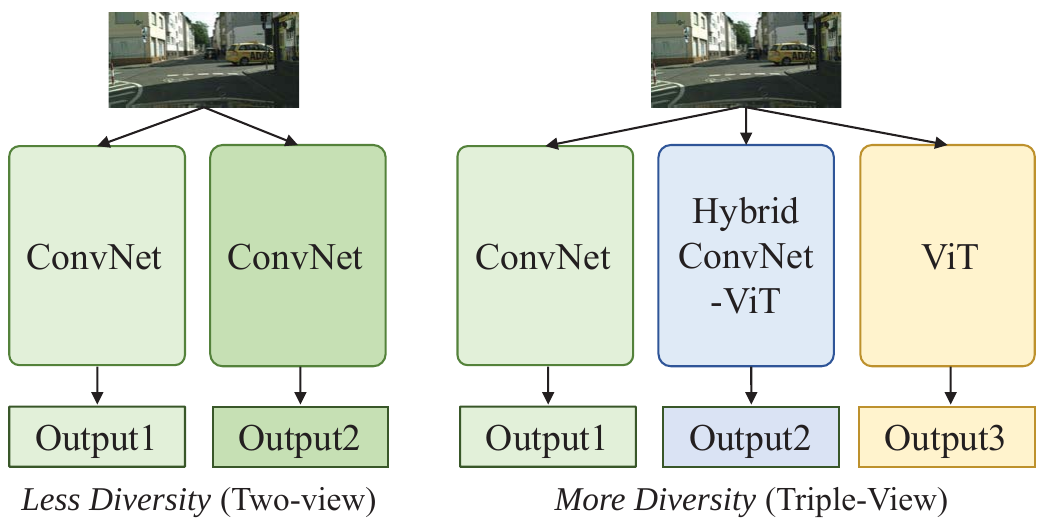}
	\caption{Motivation of Tri-KD. Left: two-view encoder with the same backbone; right: triple-view encoder (ours) using three kinds of backbones. The outputs of the former tend to be less diverse than that of the latter.}
	\label{fig:motivation}
\end{figure}

Therefore, to enhance the feature diversity and enlarge the receptive field, as depicted in Fig.~\ref{fig:motivation}, we adopt the tri-training strategy and introduce the triple-view neural networks as the encoder, which consists of pure ConvNet \cite{he-cvpr2016-resnet}, pure Vision Transformer (ViT) \cite{dosovitskiy-iclr2021-vit}, and hybrid ConvNet-ViT \cite{wu-eccv2022-tinyvit}. Among them, pure ConvNet reveals the image details by capturing the local spatial structure, facilitating the segmentation of those small or tiny objects; pure ViT models the global semantic context of image by employing the long-range pixel-level dependency; hybrid ConvNet-ViT takes advantage of the two complementary structures that encoding both the local and the global spatial relations of image pixels. Meanwhile, we adopt the knowledge distillation \cite{hinton-arxiv2015-kd} skill to transfer the knowledge from ConvNet and ViT to hybrid ConvNet-ViT by feature learning, \ie, empowering low-level features with the ability of reflecting the local details and high-level features capturing the global context of image. During inference, we only use the hybrid ConvNet-ViT as the encoder, which saves lots of computational sources allowing to be easily deployed. 

Moreover, existing semi-supervised semantic segmentation methods often use the fully-supervised model \cite{zhao-cvpr2017-pspnet,chen-pami2017-deeplabv2,chen-eccv2018-deeplabv3+} for training, but they desire expensive computational overheads. In addition, many of them \cite{wang-cvpr2022-u2pl, chen-eccv2018-deeplabv3+} use U-Net \cite{ronneberger-miccai2015-unet} or Feature Pyramid Network (FPN) \cite{lin-cvpr2017-fpn} as the decoder. For example, U-Net based methods simply concatenate the features from both encoder and decoder, but it possibly suffers from feature redundancy that takes up large memory and has adverse effect on segmentation. This inspires the FPN based methods to do the $1\times 1$ 2D convolution on the encoding features to reduce the redundancy by dimensionality reduction. However, such convolution operation may cause the missing of important spatial information, such as low-level details and high-level semantics, since it regards the features across all channels equally when reducing feature dimension. 

To overcome this shortcoming, we develop the Dual-Frequency (DF) decoder, which models the importance of encoding features before dimensionality reduction. In particular, DF decoder projects the features from the spatial domain to the frequency domain by the fast Fourier transform \cite{cochran-1967-fft}, and the feature importance is modeled by computing the channel-wise confidence scores of the dual encoding features. It utilizes the Gaussian high or low pass filter to drop out those less informative features. By this means, the model keeps those more contributing features and thus abandons the inferior features, allowing the model to be more lightweight. 

\begin{figure}[!t]
	\centering
	\includegraphics[width=\linewidth]{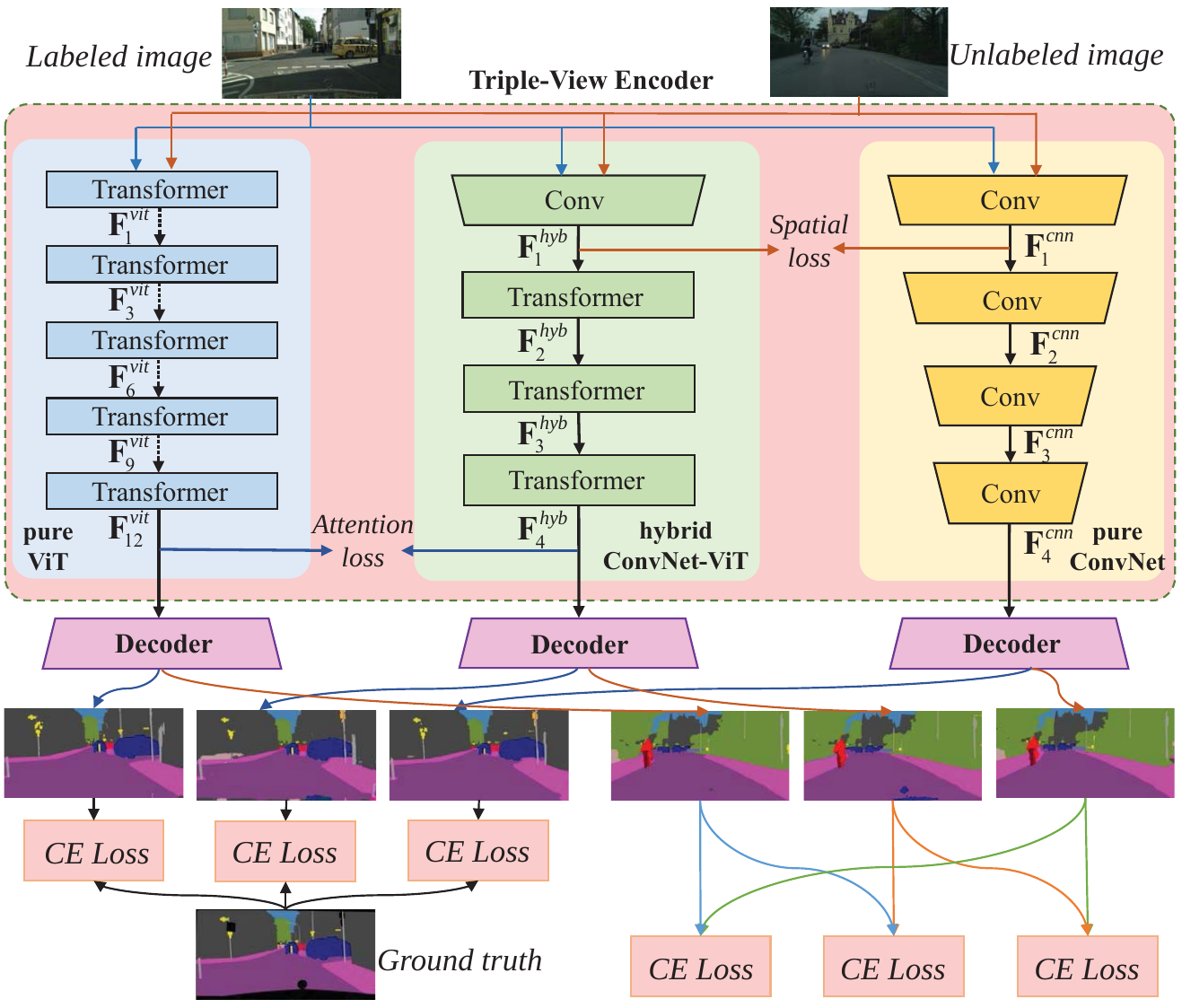}
	\caption{Overall framework of the Triple-view Knowledge Distillation (TriKD) framework for semi-supervised semantic segmentation. ``CE'' denotes cross-entropy loss, ``decoder'' adopts the dual-frequency strategy.}
	\label{fig:framework}
\end{figure}

Therefore, this work presents a Triple-view Knowledge Distillation (termed \textbf{TriKD}) framework for semi-supervised semantic segmentation, as shown in Fig.~\ref{fig:framework}. Specifically, TriKD mainly consists of the triple-view encoder and the dual-frequency decoder.
 
The main contributions are summarized as follows:
\begin{itemize}
  \item A triple-view knowledge distillation framework with dual-frequency decoding is developed for semi-supervised semantic segmentation, by considering both local spatial details and the global context of images.
  
  \item The triple-view encoder employs pure ConvNet, pure ViT, and hybrid ConvNet-ViT to learn both local and global features. Besides, the knowledge distillation strategy is applied to transfer the complementary semantics from the former two modules to hybrid ConvNet-ViT.
  
  \item The dual-frequency channel attention mechanism is used to model the importance of both the low and the high frequency features by using fast Fourier transform with Gaussian low/high pass filter.
  
  \item Empirical studies were conducted on two benchmarks, including Pascal VOC 2012 dataset~\cite{everingham-ijcv2015-pascal} and Cityscapes~\cite{cordts-cvpr2016-cityscapes}. Both quantitative and qualitative results have verified the promising segmentation performance of the proposed framework with much less inference time.

\end{itemize}

The rest of this paper is organized as follows. Section~\ref{related} reviews some closely related works and Section~\ref{method} introduces our semi-supervised semantic segmentation framework. Then, we report the experimental results on two benchmarks to demonstrate the advantage of our method in Section~\ref{test}. Finally, we conclude this work in Section~\ref{conclusion}.
%

\section{Related Work}
\label{related}
Recent years have witnessed remarkable progress in semantic segmentation. Here, we briefly discuss two settings, involving \textit{supervised} and \textit{semi-supervised} scenarios.

\subsection{Supervised Semantic Segmentation} 
Most supervised semantic segmentation works adopt deep learning models, and the milestone is the Fully-Convolution Network (FCN) \cite{long-cvpr2015-fcn} which is applied to semantic segmentation and outperforms the traditional methods. In recent, many variants of FCN have emerged, \eg, Chen \etal~\cite{chen-pami2017-deeplabv2, chen-arxiv2017-deeplabv3} present the Atrous Spatial Pyramid Pooling (ASPP) module to enlarge the receptive field of convolutional layer; Zhao \etal~\cite{zhao-cvpr2017-pspnet} devise the pyramid pooling module to segment the objects with various sizes. While they can capture multi-scale context, it is still hard to model the global semantics as convolution layers often reveal the local spatial patterns of the data. This motivates Wang \etal~\cite{wang-cvpr2018-nonlocal} to develop the non-local block using self-attention mechanism~\cite{vaswani-nips2017-attention} to model the long-range context among image pixels.

Due to the inductive bias in CNN \cite{dosovitskiy-iclr2021-vit,zheng-cvpr2021-setr}, recent works tend to adopt the sequence-to-sequence models to capture the long-range dependency for image, such as Zheng \etal~\cite{zheng-cvpr2021-setr} employ vanilla ViT \cite{dosovitskiy-iclr2021-vit} as encoder to learn multi-level features, which are upsampled progressively with feature aggregation by decoder; Wang \etal~\cite{wang-aaai2022-uctransnet} use the channel-wise cross fusion transformer to model the multi-scale context, which reduces the semantic gap between encoder and decoder. However, the above methods require very expensive human labeling costs since all pixels of training images are to be annotated in fully-supervised semantic segmentation, \eg, finer-labeling one image in Cityscapes~\cite{cordts-cvpr2016-cityscapes} database costs 1.5 hours on average. 

\subsection{Semi-supervised Semantic Segmentation} 
In semi-supervised setting, there are only a few labeled samples and a large number of unlabeled samples for semantic segmentation. Existing methods are generally divided into two groups, including \textit{self-training} and \textit{consistency regularization}.

\textbf{Self-training}. It expands the training set by generating pseudo-labeled samples. In particular, the self-training methods use the labeled samples to train a teacher network, which is employed to yield the pseudo-labels for unlabeled samples; the pseudo-labeled samples are added to the training set to learn a student network. For example, Yuan \etal~\cite{yuan-iccv2021-simplebaseline} propose the distribution-specific batch normalization to alleviate the statistical bias problem due to the large distribution difference incurred by strong data augmentation; Yang \etal~\cite{yang-cvpr2022-st++} apply the strong data augmentations on unlabeled samples to circumvent the overfitting issue of noisy labels as well as decoupling similar predictions between teacher network and student network, and performed re-training via giving priority to reliable unlabeled samples with high holistic prediction stability; Wang \etal~\cite{wang-cvpr2022-u2pl} argue that each pixel matters to the model, and employ the prediction entropy to divide the pixels to reliable and unreliable sets, which are both used to train the model; Ke \etal~\cite{ke-tip2022-three} adopts the three-stage solution to extracting pseudo-mask information on unlabeled data and enforces segmentation consistency in a multi-task fashion. However, the self-training methods iteratively learn the model by itself, and the possible noisy pseudo-labeled training samples make it difficult to obtain a good student network.

\textbf{Consistency regularization}. It applies various perturbations to the model during training and constrains the outputs with different perturbations to be consistent, such that within-class samples are closer and between-class samples are pushed far away, avoiding the overfitting to some degree. Generally, the typical perturbations are categories into image-level, feature-level, and network-level types. For image-level type, Zou \etal~\cite{zou-iclr2020-pseudoseg} do both strong and weak data augmentation on images and then feed them to the same network, and the output of the weak branch is used to supervise the strong branch since training the weak branch is more stable; French \etal~\cite{french-bmvc2020-cutmix} adopt the CutMix data augmentation to combine two images into one by rectangular mask and add them to the training set. For feature-level type, Ouali \etal~\cite{ouali-cvpr2020-cct} present the Cross-Consistency Training (CCT) strategy, which trains the model using the labeled data by the primary encoder, and then feeds the encoding features with various perturbations to corresponding decoders, whose outputs are imposed with the consistency constraint. For network-level type, Tarvainen \etal~\cite{tarvainen-nips2017-mean} obtain two augmented images by adding Gaussian noise, and one image is directly input to the model and update network parameters by using the Exponential Moving Average (EMA) skill, leading to the EMA network; the other image is fed into EMA network and its output is treated as the target of the original network output, while the consistency constraint is imposed on the two outputs by the least squares error loss. In addition, Ke \etal~\cite{ke-eccv2020-gct} propose the Guided Collaborative Training (GCT) method which feed one image to two networks with the same architecture but different initialization to yield two outputs, and one supervises the other. Furthermore, Chen \etal~\cite{chen-cvpr2021-cps} improve GCT by presenting the Cross Pseudo Supervision (CPS) approach, which adopts the one supervises the other strategy with the cross-entropy loss as the consistency constraint; Fan \etal~\cite{fan-tip2023-cpcl} train two paralleled networks via the intersection supervision using high-quality pseudo labels and union supervision using large-quantity pseudo labels; Wang \etal~\cite{wang-cvpr2023-ccvc} adopt a two-branch co-training framework that enforces two sub-nets to learn distinct features by a discrepancy loss. Nevertheless, many previous methods consider two-view co-training with the same backbone, leading to less diverse features, and fail to select those important features at the decoding stage, leading to large computational costs and lower inference speed.

\section{The Proposed Method}
\label{method}
This section describes the proposed Triple-view Knowledge Distillation (TriKD) framework as shown in Fig.~\ref{fig:framework}, which consists of triple-view encoder and dual-frequency decoder. Triple-view encoder includes pure ConvNet, pure ViT, and hybrid ConvNet-ViT, where the former two act as teacher networks and the last one is student network for knowledge distillation during training. Dual-frequency decoder adopts the channel attention to select those important features in both the low and the high frequency domain. During inference, only the hybrid ConvNet-ViT is used for encoding features, so as to speed up segmentation.

\subsection{Problem Definition}
Semi-supervised semantic segmentation task gives the pixel-level predictions by employing $N_l$ labeled images $\mathcal{D}_{l}=\{(\mathbf{X}^{l}_{i}, \mathbf{Y}^l_i)\}^{N_{l}}_{i=1}$ and $N_u$ unlabeled images $\mathcal{D}_{u}=\{\mathbf{X}^{u}_{j}\}^{N_{u}}_{j=1}$, where $N_l \ll N_u$, $\{\mathbf{X}_i^l, \mathbf{X}_j^u\} \in \mathbb{R}^{H\times W\times 3}$ denotes the $i$-th labeled and the $j$-th unlabeled RGB image, respectively; $H$ denotes the height and $W$ denotes the width; $\mathbf{Y}_i^l \in \mathbb{R}^{H\times W\times C}$ is the ground-truth label map of the $i$-th image and $C$ is the class number. For brevity, we omit the subscripts, \ie, $\{\mathbf{X}^l, \mathbf{Y}^l, \mathbf{X}^u\}$ denote one labeled image and its label map, and one unlabeled image, respectively. 

\subsection{Triple-View Encoder}
To learn diverse and complementary features, we employ the tri-training strategy \cite{dong-ijcai2018-trinet} to train the segmentation model by using three different backbones for feature encoding. In particular, pure ConvNet (ResNet \cite{he-cvpr2016-resnet}) reveals the local patterns of the data, pure ViT (DeiT \cite{touvron-icml2021-deit}) respects the global structure of the data, and hybrid ConvNet-ViT (TinyViT \cite{wu-eccv2022-tinyvit}) inherits the merits of the former two by adopting the knowledge distillation skill. For unlabeled data, we impose the consistency regularization on the predictions from the three networks. 

\textbf{ConvNet}. Due to the limited receptive field of convolution, it is common to stack multiple convolution layers and pooling layers to enlarge the receptive field. For pure ConvNet, we use ResNet \cite{he-cvpr2016-resnet} as the backbone, which has four stages with downsampling at each stage. When one image goes through ResNet, we have four feature maps $\mathbf{F}^{cnn}_s \in \mathbb{R}^{{H_s}\times{W_s}\times{C^{cnn}_s}}$, where the subscript $s\in \{1,2,3,4\}$ denotes the stage, the superscript $cnn$ denotes the ConvNet, $H_s \times W_s$ denotes the resolution of feature map (height$\times$width), $C_s^{cnn}$ denotes the channel number. Here, $\{H_1, W_1, C_1^{cnn}\}=\{\frac{H}{4}, \frac{W}{4}, 256\}$, $\{H_{s+1}, W_{s+1}, C_{s+1}^{cnn}\} = \{\frac{H_s}{2}, \frac{W_s}{2}, 2\cdot C_s^{cnn} \}$. 

\textbf{ViT}. Generally, it is crucial to capture the long-range pixel dependency to understand large objects and global scene in image. To achieve this, we adopt the light DeiT \cite{touvron-icml2021-deit} as the backbone of ViT, which contains multiple self-attention layers to model the global pixel relations. Following \cite{zheng-cvpr2021-setr}, one image is divided into many small patches with the size of $P \times P\times 3$, each of which is linearly projected to a feature embedding vector $\mathbf{e}\in \mathbb{R}^{C^{vit}}$. The number of small patches is computed by $N^{vit}=\frac{H\cdot W}{P^2}$, \ie, the length of patch sequence, and one image has $N^{vit}$ feature embedding vectors, which compose the patch matrix $\mathbf{F}_0^{vit}\in \mathbb{R}^{N^{vit}\times C^{vit}}$ as the input of ViT. Here, we empirically set $P$ to 16 and $C^{vit}$ to 768. The backbone DeiT consists of multiple encoding stages, and each stage contains a series of stacked transformer layers. The transformer layer includes Multi-head Self-Attention (MSA) \cite{zheng-cvpr2021-setr} module and Multi-Layer Perceptron (MLP) \cite{mitra-tnn1995-mlp}. At the $r$-th stage, it learns the features by computing
\begin{equation}
	\hat{\mathbf{F}}^{vit}_{r} = MSA(LN(\mathbf{F}^{vit}_{r-1})) + \mathbf{F}^{vit}_{r-1},
\end{equation}
\begin{equation}
	\mathbf{F}^{vit}_{r} = MLP(LN(\hat{\mathbf{F}}^{vit}_{r})) + \hat{\mathbf{F}}^{vit}_{r},
\end{equation}
where $r \in \left\{ 1,2,\cdots,12 \right\}$ indexes the encoding stage, $LN(\cdot)$ denotes layer normalization, and the feature size keeps the same across stages, \ie, $\mathbf{F}^{vit}_r\in \mathbb{R}^{N^{vit}\times C^{vit}}$.

\textbf{Hybrid ConvNet-ViT}. To take advantage of both ConvNet and ViT, we adopt TinyViT \cite{wu-eccv2022-tinyvit} as the backbone of hybrid ConvNet-ViT. In particular, it comprises four encoding stages, where the first stage is convolution layer and the remaining stages are transformer layers. We define the feature map of the first stage is $\mathbf{F}^{hyb}_1 \in \mathbb{R}^{H_1\times W_1\times C^{hyb}_1}$, where ``hyb'' denotes hybrid ConvNet-ViT and $C^{hyb}_1=64$ denotes the feature channel \aka~embedding dimension. The feature maps of the remaining three stages are represented by $\mathbf{F}^{hyb}_k \in \mathbb{R}^{L^{hyb}_k\times C^{hyb}_k}$, where $C^{hyb}_{k+1}=2\cdot C^{hyb}_k$, $k=1,2$ indexes the stage, $C^{hyb}_4=448$, and $L^{hyb}_k$ denotes the sequence length of feature embedding. To keep the feature size consistency with that of ResNet, we follow \cite{wu-eccv2022-tinyvit} to add the downsampling operation between pairwise stages, and do serialization to satisfy $L^{hyb}_2 = {H_2}\cdot{W_2}, L^{hyb}_3 = {H_3}\cdot{W_3}, L^{hyb}_4 = {H_4}\cdot{W_4}$.
%

\begin{figure*}[!t]
	\centering
	\includegraphics[width=0.8\linewidth]{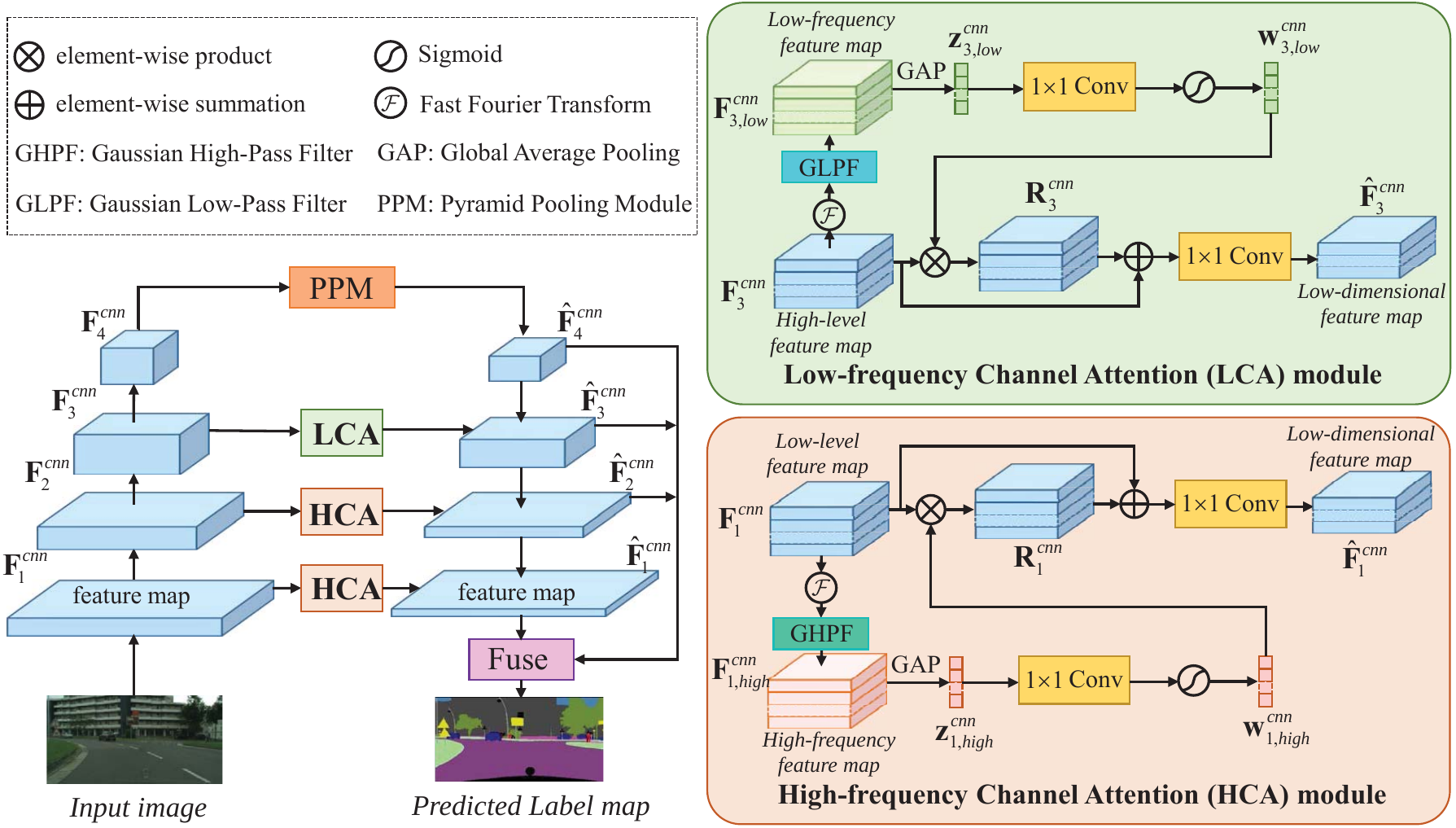}
	\caption{Architecture of dual-frequency decoder. Take pure ConvNet for example.}
	\label{fig:decoder}
\end{figure*}

\subsection{Knowledge Distillation Scheme}
We adopt the knowledge distillation \cite{hinton-arxiv2015-kd} scheme to transfer the knowledge learned from the larger network to the smaller one. In particular, we regard pure ConvNet and pure ViT as teacher network while hybrid ConvNet-ViT is taken as student network. Here the knowledge denotes the ability of capturing local spatial patterns by convolution network, and that of modeling the global long-range dependency of pixels by vision transformer. In another word, we expect hybrid ConvNet-ViT learns both the local and global pixel structures of the image from teacher networks at the cost of much smaller model size and lower computations, allowing it to be readily deployed.   

As shown in Fig.~\ref{fig:framework}, both the labeled and unlabeled images are fed into the triple-view encoder to yield three feature maps, \ie, the local feature map by pure ConvNet, the global attention map by pure ViT, and the distillation feature map by hybrid ConvNet-ViT. To learn the knowledge from teacher network, we impose the consistency constraints on the low-level feature maps by the spatial loss ($\ell_2$-loss) and on the high-level feature maps by the attention loss (Kullback-Leibler divergence), respectively. The former helps student network to reveal the local structure of images while the latter allows student network to capture semantic features by modeling the global context.

\textbf{Low-level distillation}. To capture the local spatial relations of pixels, we conduct the distillation between the low-level layers of pure ConvNet and that of student (hybrid) network. In particular, we adopt the convolution operation with size of $1\times 1$ to make the channel number of feature map $\mathbf{F}^{hyb}_1$ from student network be the same as that of pure ConvNet at the first stage, followed by batch normalization $BN(\cdot)$ and a nonlinear function, resulting in the low-level distilled feature $\tilde{\mathbf{F}}^{hyb}_1\in \mathbb{R}^{H_1\times W_1 \times C^{cnn}_1}$, \ie,
\begin{equation} 
	\tilde{\mathbf{F}}^{hyb}_1 = ReLU(BN(Conv2D(\mathbf{F}^{hyb}_1)),
\end{equation}
where $Conv2D(\cdot)$ denotes $1\times 1$ convolution, and $ReLU(\cdot)$ is Rectified Linear Unit which is an activation function. Naturally, the low-level distilled feature is expected to share with the local structure of the image encoded by pure ConvNet. This is achieved by imposing the consistency constraint on the low-level features, \ie, the spatial loss  $\mathcal{L}_{spa}$:
\begin{equation}
	\mathcal{L}_{spa} = \frac{1}{H_1\cdot W_1\cdot {C^{cnn}_1}} \sum_{c=1}^{C_1^{cnn}}
	                    \|\mathbf{F}^{cnn}_{1,c} - \tilde{\mathbf{F}}^{hyb}_{1,c}\|_F^2,
\end{equation}
where $\|\cdot\|_F$ denotes the Frobenius norm of matrix, and $c$ indexes the channel feature map $\{\tilde{\mathbf{F}}_{1,c}^{hyb},\mathbf{F}_{1,c}^{cnn} \}\in \mathbb{R}^{H_1\times W_1}$. With this constraint, the low-level hybrid layer is able to learn the spatial knowledge from the corresponding ConvNet layer.

\textbf{High-level distillation}. To capture the global relations of pixels, we conduct the distillation between the high-level layers of pure ViT and that of student (hybrid) network. In particular, we first compute their attention maps $\mathbf{A}$ by Multi-head Self-Attention \cite{zheng-cvpr2021-setr} and then impose the consistency constraint on them. Mathematically, the query $\mathbf{Q}$ and the key $\mathbf{K}$ values are computed by (subscripts are omitted for brevity)
\begin{align}
	\{\mathbf{Q}^{vit}, \mathbf{K}^{vit}\} & = \{\mathbf{F}^{vit} \mathbf{W}^{vit}_{Q},\mathbf{F}^{vit}\mathbf{W}^{vit}_{K}\} \in \mathbb{R} 	^{N^{vit}\times d^{vit}}, \\
	\{\mathbf{Q}^{hyb}, \mathbf{K}^{hyb}\} & = \{\mathbf{F}^{hyb} \mathbf{W}^{hyb}_{Q},\mathbf{F}^{hyb}\mathbf{W}^{hyb}_{K}\} \in \mathbb{R} 	^{L^{hyb}\times d^{hyb}}, 
\end{align}
where $\{\mathbf{W}^{vit}_Q,\mathbf{W}^{vit}_K\} \in \mathbb{R}^{C^{vit}\times d^{vit}}$ and  $\{\mathbf{W}^{hyb}_Q,\mathbf{W}^{hyb}_K\} \in \mathbb{R}^{C^{hyb}\times d^{hyb}}$ are learnable parameters, the head dimensions are $\{d^{vit}, d^{hyb}\}=\{\frac{C^{vit}}{m^{vit}}, \frac{C_4^{hyb}}{m^{hyb}}\}$ where $\{m^{vit}, m^{hyb}\}=\{12, 14\}$ are the head numbers.

and the corresponding attention maps are computed by
\begin{align}
\mathbf{A}^{vit} & = Softmax(\frac{\mathbf{Q}^{vit}(\mathbf{K}^{vit})^\top}{\sqrt{d^{vit}}})\in \mathbb{R} 	^{N^{vit}\times N^{vit}}, \\
\mathbf{A}^{hyb} & = Softmax(\frac{\mathbf{Q}^{hyb}(\mathbf{K}^{hyb})^\top}{\sqrt{d^{hyb}}})\in \mathbb{R} 	^{L^{hyb}\times L^{hyb}},
\end{align}
where $Softmax(\cdot)$ is the softmax function. Hopefully, the high-level distilled feature is expected to model the global structure revealed by pure ViT. This is achieved by imposing the consistency constraint on the high-level features, \ie, the attention loss  $\mathcal{L}_{att}$:
\begin{equation}
	\mathcal{L}_{att} = \frac{1}{{N^{vit}}\cdot {N^{vit}}} KL(\mathbf{A}^{vit} || Upsample(\mathbf{A}^{hyb})),
\end{equation}
where $KL(\cdot || \cdot)$ denotes the Kullback-Leibler divergence loss, and $Upsample(\cdot)$ denotes bilinear interpolation for making the dimension of $\mathbf{A}^{hyb}$ be the same as that of $\mathbf{A}^{vit}$. By this means, the high-level features can well capture the global context of the image pixels by distilling the semantics from the high-level transformer layer of ViT.

\subsection{Decoder}
To obtain the predicted label map, existing methods always employ convolution-like U-Net \cite{ronneberger-miccai2015-unet} or Feature Pyramid Network (FPN) \cite{lin-cvpr2017-fpn} as the decoder, by either concatenating the features of encoder and decoder along the channel dimension or reducing the feature dimension via $1\times 1$ convolution for abandoning the redundancy. Thus, all features are equally treated during the decoding period, which may possibly include some redundancy content. This results in the failure of well capturing the structural context of the image, such as the low-level details and the high-level semantics. To address this problem, we model the hierarchical feature importance by developing the dual-frequency channel attention mechanism for a FPN-like decoder based on UPerNet (Unified Perceptual Parsing Network) \cite{xiao-eccv2018-upernet}, and we shortly call it dual-frequency decoder. 

Essentially, the idea of devising the dual-frequency decoder originates from the fact that low-frequency signal describes the main part of an image and high-frequency signal describes its details. According to \cite{magid-iccv2021-dynamic}, neural networks tend to learn low-frequency signal while neglecting high-frequency signal, and low-level features involve rich high-frequency signals while high-level features involve plentiful low-frequency signals. Therefore, inspired by \cite{li-cvpr2020-wavelet}, we consider the channel importance of hierarchical features by projecting the features from the spatial domain to the frequency domain, and obtain the enhanced features by channel-wise attention mechanism for decoding.

Here are the details of decoder. As shown in Fig.~\ref{fig:decoder}, it has four stages and involves Pyramid Pooling Module (PPM) \cite{zhao-cvpr2017-pspnet}, multi-scale feature Fusion module (Fuse), Low-frequency Channel Attention module (LCA), and High-frequency Channel Attention module (HCA). In particular, PPM accepts the features of the last encoding stage, while LCA and HCA are applied to capture the global context and the local details of images before using $1\times 1$ convolution to obtain the low-dimensional features. The hierarchical feature maps of the four decoding stages are fused by multi-scale feature fusion module, resulting in the fused feature map, which generates the predicted label map after upsampling. 

For our triple-view encoder, it desires some feature processing before passing through the decoder. Specifically, the four-stage features of pure ConvNet can be directly used; the features at the last three encoding stages of hybrid network undergo the deserialization \cite{wu-eccv2022-tinyvit}; the features at the third, sixth, ninth, 12-th transformer layers of pure ViT also undergo the deserialization, \ie, reshape 2D feature to 3D one with a size of $\frac{H}{P}\times \frac{W}{P}\times C^{vit}$, which are then passed to $1\times 1$ Conv and $3\times 3$ Conv as well as upsampling. 

Note that we keep the fourth stage that uses PPM to capture the rich context of high-level feature by the feature map $\hat{\mathbf{F}}_4 \in \mathbb{R}^{H_4\times W_4\times C_4}$ (superscripts omitted) and do dimensionality reduction on the features of the remaining stages. As shown in Fig.~\ref{fig:decoder}, the low-level features of the first and the second stages, \ie, $\{\mathbf{F}_1^{cnn},\mathbf{F}_2^{cnn}\}$, are passed to the HFA module for modeling the global context, while the high-level features of the third stage, \ie, $\mathbf{F}_3^{cnn}$, are passed to the LFA module for modeling the local details. This requires first projecting the features from the spatial domain to the frequency domain using Fast Fourier Transform (FFT) \cite{cochran-1967-fft}, and then using $fftshit(\cdot)$ in PyTorch to move the low or high frequency signal to the middle, leading to the spectral map. These spectral maps pass through Gaussian Low-Pass or High-Pass Filter (GLPF or GHPF) \cite{deng-1993-gaussian} to obtain the low-frequency feature maps $\mathbf{F}_{3, low}^{cnn}$ and the high-frequency feature maps $\{\mathbf{F}_{1,high}^{cnn}, \mathbf{F}_{2,high}^{cnn}\}$. Thus, the high-frequency signals in the low-level feature map keep the image details and the low-frequency signals in the high-level feature map carry the image context on a whole. Thereafter, we do the Global Average Pooling (GAP) operation on the high-frequency and the low-frequency signals to obtain the corresponding vectors $\{\mathbf{z}_{1,high}^{cnn}\in \mathbb{R}^{C_1^{cnn}}, \mathbf{z}_{2,high}^{cnn}\in \mathbb{R}^{C_2^{cnn}}, \mathbf{z}_{3,low}^{cnn}\in \mathbb{R}^{C_3^{cnn}}\}$, for each of which the length equals the channel number. We model the channel importance by applying the $1\times 1$ 2D convolution and the sigmoid function $\sigma(\cdot)$ to the above vectors, whose entries acts as the importance scores of different channels. Hence, we can obtain the weighted feature map $\mathbf{R}^{cnn}$, \ie,
\begin{equation}
	\mathbf{R}^{cnn}=\sigma(Conv2D(\mathbf{z}^{cnn})) \otimes \mathbf{F}^{cnn} \in \mathbb{R}^{H\times W\times C^{cnn}},
\end{equation}
where the subscripts are omitted, and the important channels have larger weights than those less important ones in the feature map. Later, the low-level and the high-level feature maps are concatenated with the corresponding weighted feature maps by the skip connection. At last, we use the $1\times 1$ 2D convolution to reduce the channel dimension, resulting in the low-dimensional feature representations $\{\hat{\mathbf{F}}_1^{cnn}\in \mathbb{R}^{H_1\times W_1\times \hat{C}_1^{cnn}}, \hat{\mathbf{F}}_2^{cnn}\in \mathbb{R}^{H_2\times W_2\times \hat{C}_2^{cnn}}, \hat{\mathbf{F}}_3^{cnn}\in \mathbb{R}^{H_3\times W_3\times \hat{C}_3^{cnn}}\}$, where $\{\hat{C}_1^{cnn},\hat{C}_2^{cnn},\hat{C}_3^{cnn}\}=\{128,256,256\}$. During the multi-scale feature fusion, we apply the $1\times 1$ 2D convolution to match the dimension of the previous feature map with that of the current one. 

Similar procedures are applied for the pure ViT and the student network, where $\{\hat{C}_1^{vit},\hat{C}_2^{vit},\hat{C}_3^{vit}\}= \{\hat{C}_1^{hyb},\hat{C}_2^{hyb},\hat{C}_3^{hyb} \}=\{48, 96, 192\}$. In this way, we obtain the predicted label maps $\{\hat{\mathbf{Y}}^{l,cnn}, \hat{\mathbf{Y}}^{l,vit}, \hat{\mathbf{Y}}^{l,hyb}\}\in \mathbb{R}^{H\times W\times C}$ and $\{\hat{\mathbf{Y}}^{u,cnn}, \hat{\mathbf{Y}}^{u,vit}, \hat{\mathbf{Y}}^{u,hyb}\} \in \mathbb{R}^{H\times W\times C}$ for the labeled images $\mathbf{X}^l$ and the unlabeled images $\mathbf{X}^u$, respectively.

\subsection{Loss Function}
To optimize the semi-supervised segmentation model, we adopt three kinds of losses, including the segmentation loss, the distillation loss, and the Cross Pseudo Supervision (CPS) \cite{chen-cvpr2021-cps} loss. Among them, the segmentation loss is applied to labeled samples, the CPS loss is applied to unlabeled samples, and the distillation loss is applied to all training samples.

\textbf{Segmentation loss}. For labeled images, we adopt the Cross-Entropy (CE) loss to compute the segmentation loss of the predicted label maps and ground-truth ones, \ie,
\begin{equation}
	\begin{split}
		\{\mathcal{L}_{seg}^{cnn},\mathcal{L}_{seg}^{vit},\mathcal{L}_{seg}^{hyb}\} & = -\frac{1}{N_l\cdot HW} \cdot \\
		~  \Big\{ \sum_{i=1}^{N_l}\mathbf{Y}_i^l \log \hat{\mathbf{Y}}_i^{l,cnn}, &
		\sum_{i=1}^{N_l}\mathbf{Y}_i^l \log \hat{\mathbf{Y}}_i^{l,vit}, 
		\sum_{i=1}^{N_l}\mathbf{Y}_i^l \log \hat{\mathbf{Y}}_i^{l,hyb} ~\Big\},
	\end{split}
\end{equation}
where $HW=H\cdot W$ and $\mathcal{L}_{seg} = \mathcal{L}_{seg}^{cnn} + \mathcal{L}_{seg}^{vit} +\mathcal{L}_{seg}^{hyb}$.

\textbf{CPS loss}. For unlabeled images, we adopt the CPE loss to make the predicted label maps be consistent with different encoders. In particular, we apply the argmax function to the predicted label map to obtain the one-hot label map of one encoder as the supervision of another encoder, and adopt the CE loss to compute the cross pseudo supervision loss below:
\begin{align*}
	\mathcal{L}_{cps}^{cnn} &=   -\frac{1}{N_u\cdot HW} \cdot \sum_{j=1}^{N_u} 
	(\hat{\mathbf{Y}}_j^{u, vit} + \hat{\mathbf{Y}}_j^{u, hyb}) \cdot \log \hat{\mathbf{Y}}_j^{u,cnn}, \\
	\mathcal{L}_{cps}^{vit} &=   -\frac{1}{N_u\cdot HW} \cdot \sum_{j=1}^{N_u} 
	(\hat{\mathbf{Y}}_j^{u, cnn} + \hat{\mathbf{Y}}_j^{u, hyb}) \cdot \log \hat{\mathbf{Y}}_j^{u,vit}, \\
	\mathcal{L}_{cps}^{hyb} &=   -\frac{1}{N_u\cdot HW} \cdot \sum_{j=1}^{N_u} 
	(\hat{\mathbf{Y}}_j^{u, cnn} + \hat{\mathbf{Y}}_j^{u, vit}) \cdot \log \hat{\mathbf{Y}}_j^{u,hyb}, 
\end{align*}
and $\mathcal{L}_{cps} = \mathcal{L}_{cps}^{cnn} + \mathcal{L}_{cps}^{vit} +\mathcal{L}_{cps}^{hyb}$.

\textbf{Distillation loss}. For all training samples, we use the sum of the spatial loss $\mathcal{L}_{spa}$ and the attention loss $\mathcal{L}_{att}$ as the distillation loss, such that the student network inherits the merits of the teacher networks including modeling the locality property by pure ConvNet and the global context by pure ViT, \ie, $\mathcal{L}_{kd} = \lambda_1\mathcal{L}_{spa} + \lambda_2\mathcal{L}_{att}$, where the regularization constants are equally set to 0.5.

\textbf{Total loss}. To optimize the objective of our Triple-view Knowledge Distillation (TriKD) framework, we compute the total loss as follows:
\begin{equation}
	\mathcal{L} = \mathcal{L}_{seg} + \mathcal{L}_{kd} + \lambda\mathcal{L}_{cps},
\end{equation}
where the constant $\lambda$ is empirically set to 0.1.


\section{Experiment}
\label{test}
This section shows extensive experiments of semantic segmentation on two benchmark data sets. All experiments were conducted on a machine with four NVIDIA RTX 3090 graphics cards, and our model was compiled using PyTorch 1.10.0, Python 3.9, and CUDA 11.4.

\subsection{Datasets and Evaluation Metrics}
\textbf{Pascal VOC 2012}\footnote{\url{http://host.robots.ox.ac.uk/pascal/VOC/voc2012/}}~\cite{everingham-ijcv2015-pascal}. It has 1464 training images, 1449 validation images, and 1456 test images, which involve 20 object categories and one background class. Following \cite{chen-cvpr2021-cps,wang-cvpr2022-u2pl}, we add the 9118 training images of the Segmentation Boundary Database \cite{hariharan-iccv2011-semantic} to augment the training set as ``VOCAug''. 

\textbf{Cityscapes}\footnote{\url{https://www.cityscapes-dataset.com/}}~\cite{cordts-cvpr2016-cityscapes}. It focuses on semantic understanding of urban street scenes, and the images are taken from 50 cities by several months in daytime with good/medium weather conditions. Each annotated image is the 20th image from a 30 frame video snippets (1.8s), and there are 5000 annotated images with high-quality dense pixel annotations involving 10 object classes. This dataset has 2975 training images, 500 validation images, and 1525 test images. 

\textbf{Evaluation Metrics}. Following \cite{chen-cvpr2021-cps,wang-cvpr2022-u2pl}, we adopt the commonly used metric, \ie, mIoU (mean Intersection over Union), to evaluate the semantic segmentation performance of the compared methods. It is the average IoU score of the predicted and the ground-truth semantic region across all classes. Note that empirical results were obtained on the validation set, and ablation studies were conducted on VOCAug and Cityscapes.

\subsection{Experimental Settings}
\textbf{Backbone}. Our TriKD framework adopts the triple-view encoder that employs ResNet101 \cite{he-cvpr2016-resnet} for pure ConvNet, DeiT-B \cite{touvron-icml2021-deit} for pure ViT, and TinyViT \cite{wu-eccv2022-tinyvit} for hybrid ConvNet-ViT. To show precision-speed trade off, we use two variants of TinyViT, \ie, TinyViT-11M for TriKD and TinyViT-21M for TriKD$^{\ast}$ with different model sizes.

\textbf{Training Phase}. All encoders initialize with the weights of the models pre-trained on the ImageNet \cite{deng-cvpr2009-imagenet} dataset, and the weights of other layers use the Kaiming initialization \cite{he-iccv2015-delving} in PyTorch. The initial learning rate $\ell_r$ is set to 0.001 for VOC and 0.005 for Cityscapes with the polynomial learning rate decay, where $\ell_r$ is multiplied by $(1 - \frac{iter}{iter_{max}})^{power}$ with $power=0.9$. We adopt the stochastic gradient descent algorithm to update model parameters, the momentum is set to 0.9, and the weight decay is set to 0.0001 for VOC with 80 epochs and 0.0005 for Cityscapes with 300 epochs. The batch size is 8, which decides the maximum iteration with the epochs. For data augmentation, we adopt random horizontal flipping, randomly scaling images by 0.5 to 2 factors, random cropping by $515\times 512$, and random rotation in $[-10^\circ, 10^\circ]$. The pixel values of all images are normalized to $[0, 1]$. 

\textbf{Inference Phase}. We normalize the test image and feed it to the hybrid ConvNet-ViT of the trained model, which yields the predicted label map $\hat{\mathbf{Y}}$ by the corresponding decoder.

\subsection{Compared Methods}
We compare two groups of state-of-the-art semi-supervised semantic segmentation methods, \ie, 1) \emph{self-training} group including SDA (Strong Data Augmentation) \cite{yuan-iccv2021-simplebaseline}, ST++ (advanced Self-Training) \cite{yang-cvpr2022-st++}, U$^2$PL (Using Unreliable Pseudo-Labels) \cite{wang-cvpr2022-u2pl}, and TSS (Three-Stage Self-training) \cite{ke-tip2022-three}; 2) \emph{consistency regularization} group, including image-level perturbation methods PseudoSeg \cite{zou-iclr2020-pseudoseg} and CutMix \cite{french-bmvc2020-cutmix}, feature-level perturbation method CCT (Cross-Consistency Training) \cite{ouali-cvpr2020-cct}, and network-level perturbation methods MT (Mean Teacher) \cite{tarvainen-nips2017-mean}, GCT (Guided Collaborative Training) \cite{ke-eccv2020-gct}, CPS (Cross Pseudo Supervision) \cite{chen-cvpr2021-cps}, CPCL (Conservative Progressive Collaborative Learning) \cite{fan-tip2023-cpcl}, and CCVC (Conflict-based Cross-View Consistency) \cite{wang-cvpr2023-ccvc}. Our Tri-KD belongs to the latter group and has four versions, including vanilla TriKD and supervised TriKD (TriKD$_s$, S-11M) with TinyViT-11M, as well as the two (TriKD$^\ast$, TriKD$_s^\ast$, S-21M) with TinyViT-21M, among which the supervised version means only using labeled samples for training.

\begin{table}[!t]
	\centering
	\caption{Performance comparison (mIoU \%) on Pascal VOC \cite{everingham-ijcv2015-pascal}. }
	\label{tbl:result_voc}
	\begin{tabular}{l c c c c c c}
		\toprule[0.75pt]
		\multirow{2}{*}{Methods} & \multirow{2}{*}{Venue} & \multicolumn{1}{c}{1/16} & \multicolumn{1}{c}{1/8} & \multicolumn{1}{c}{1/4} & \multicolumn{1}{c}{1/2} & \multicolumn{1}{c}{Full}	\\
		\cmidrule[0.5pt]{3-7} %
		&   & 92 & 183 & 366 & 732 & 1464 \\
		\midrule[0.5pt]
		CutMix\cite{french-bmvc2020-cutmix} & BMVC'20 & 52.16 & 63.47 & 69.46 & 73.73 & 76.54 \\
		PseudoSeg\cite{zou-iclr2020-pseudoseg} & ICLR'21 & 57.60 & 65.50 & 69.14 & 72.41 & 73.23 \\
		SDA\cite{yuan-iccv2021-simplebaseline} & ICCV'21 & - & - & - & - & 75.00\\
		ST++\cite{yang-cvpr2022-st++} & CVPR'22 & 65.20 & 71.00 & 74.60 & 77.30 & 79.10 \\
		U$^{2}$PL\cite{wang-cvpr2022-u2pl} & CVPR'22 & 67.98 & 69.15 & 73.66 & 76.16 & 79.49 \\
		CCVC\cite{wang-cvpr2023-ccvc} & CVPR'23 & \underline{70.20} & \textbf{74.40} & \textbf{77.40} & \underline{79.10} & \underline{80.50} \\
		\midrule[0.5pt]
		TriKD$_s$ & S-11M & 44.32 & 58.84 & 67.40 & 72.94 & 75.67 \\
		TriKD & 11M & 67.00 & 68.54 & 74.04 & 76.52 & 79.71 \\
		TriKD$_s^\ast$ & S-21M & 55.22 & 64.84 & 70.45 & 75.01 & 78.63 \\
		TriKD$^\ast$ & 21M &  \textbf{70.29} & \underline{72.53} & \underline{76.80} & \textbf{79.71} & \textbf{81.13} \\
		\toprule[0.75pt]
	\end{tabular}
\end{table}

\begin{table}[!t]
	\centering
	\caption{Performance comparison (mIoU \%) on VOCAug \cite{everingham-ijcv2015-pascal}. }
	\label{tbl:result_vocaug}
	\begin{tabular}{l c c c c c}
		\toprule[0.75pt]
		\multirow{2}{*}{Methods} & \multirow{2}{*}{Venue} & \multicolumn{1}{c}{1/16} & \multicolumn{1}{c}{1/8} & \multicolumn{1}{c}{1/4} & \multicolumn{1}{c}{1/2} \\
		\cmidrule[0.5pt]{3-6} %
		&   & 662 & 1323 & 2646 & 5291 \\	
		\midrule[0.5pt]
		MT\cite{tarvainen-nips2017-mean} & NeurIPS'17 & 70.51 & 71.53 & 73.02 & 76.58 \\
		CCT\cite{ouali-cvpr2020-cct} & CVPR'20  & 71.86 & 73.68 & 76.51 & 77.40 \\
		CutMix-Seg\cite{french-bmvc2020-cutmix} & BMVC'20 & 71.66 & 75.51 & 77.33 & 78.21 \\
		GCT\cite{ke-eccv2020-gct} & ECCV'20 & 70.90 & 73.29 & 76.66 & 77.98 \\
		CPS\cite{chen-cvpr2021-cps} & CVPR'21 & 74.48 & 76.44 & 77.68 & 78.64 \\
		ST++\cite{yang-cvpr2022-st++} & CVPR'22 & 74.70 & 77.90 & 77.90 & - \\
		U$^{2}$PL\cite{wang-cvpr2022-u2pl} & CVPR'22 & 77.21 & 79.01 & 79.30 & \underline{80.50}\\
		TSS\cite{ke-tip2022-three} & TIP'22 & - & 72.95 & - &  - \\
		CPCL\cite{fan-tip2023-cpcl} & TIP'23 & 71.66 & 73.74 & 74.58 & 75.30 \\
		CCVC\cite{wang-cvpr2023-ccvc} & CVPR'23 & 77.20 & 78.40 & 79.00 & - \\
		\midrule[0.5pt]
		TriKD$_s$ & S-11M & 70.31 & 72.80 & 75.66 & 77.91 \\
		TriKD     &  11M & \underline{77.45} & \underline{79.23} & \underline{79.52} & 80.03 \\
		TriKD$_s^\ast$ & S-21M & 72.52 & 74.03 & 77.63 & 79.52 \\
		TriKD$^\ast$ &  21M & \textbf{78.75} & \textbf{80.42} & \textbf{81.83} & \textbf{82.27} \\
		\toprule[0.75pt]
	\end{tabular}
\end{table}

\begin{table}[!t]
	\centering
	\caption{Performance comparison (mIoU \%) on Cityscapes \cite{cordts-cvpr2016-cityscapes}. }
	\label{tbl:result_cityscapes}
	\begin{tabular}{l c c c c c}   
		\toprule[0.75pt]
		\multirow{2}{*}{Methods} & \multirow{2}{*}{Venue} & \multicolumn{1}{c}{1/16} & \multicolumn{1}{c}{1/8} & \multicolumn{1}{c}{1/4} & \multicolumn{1}{c}{1/2} \\
		\cmidrule[0.5pt]{3-6} %
		&   & 186 & 372 & 744 & 1488 \\	
		\midrule[0.5pt]
		MT\cite{tarvainen-nips2017-mean} & NeurIPS'17 & 69.03 & 72.06 & 74.20 & 78.15 \\
		CCT\cite{ouali-cvpr2020-cct} & CVPR'20 &  69.32 & 74.12 & 75.99 & 78.10 \\
		CutMix\cite{french-bmvc2020-cutmix} & BMVC'20 & 67.06 & 71.83 & 76.36 & 78.25 \\
		GCT\cite{ke-eccv2020-gct} & ECCV'20 &  66.75 & 72.66 & 76.11 & 78.34 \\
		CPS\cite{chen-cvpr2021-cps} & CVPR'21 & 69.78 & 74.31 & 74.58 & 76.81 \\
		SDA\cite{yuan-iccv2021-simplebaseline} & ICCV'21 & -	 & 74.10 & \underline{77.80} & 78.70 \\
		ST++\cite{yang-cvpr2022-st++} & CVPR'22 & -	 & 72.70 & 73.80 & - \\
		U$^{2}$PL\cite{wang-cvpr2022-u2pl} & CVPR'22 & 70.30 & 74.37 & 76.47 & \underline{79.05} \\
		TSS\cite{ke-tip2022-three} & TIP'22 & - & 62.82 & 65.80 & 67.11 \\
		CPCL\cite{fan-tip2023-cpcl} & TIP'23 & 69.92 & 74.60 & 76.98 & 78.17 \\
		CCVC\cite{wang-cvpr2023-ccvc} & CVPR'23 & \textbf{74.90} & \underline{76.40} & 77.30 & -\\
		\midrule[0.5pt]
		TriKD$_s$ & S-11M & 62.51 & 68.07 & 70.25 & 72.96 \\
		TriKD  & 11M & 69.18 & 73.25 & 75.02 & 75.98 \\
		TriKD$_s^\ast$ & S-21M & 67.06 & 72.40 & 73.33 & 75.95 \\
		TriKD$^\ast$ & 21M& \underline{72.70} & \textbf{76.44} & \textbf{78.01} & \textbf{79.12} \\
		\toprule[0.75pt]
	\end{tabular}
\end{table}

\begin{table}[!t]
	\centering
	\caption{Comparison of model parameters and inference speed. Encoders all adopts ResNet101 except ours, ``DL'' is DeepLab, ``DF'' is dual-frequency decoder.}
	\label{tbl:speed_modelPara}
	\setlength{\tabcolsep}{0.6mm}{  
		\begin{tabular}{l c c c c  c  c c c}
			\toprule[0.75pt]
			\multirow{2}{*}{Method} & \multirow{2}{*}{Venue} &  \multirow{2}{*}{Decoder} & \multicolumn{1}{c}{Params} & \multicolumn{2}{c}{VOC \cite{everingham-ijcv2015-pascal}} && \multicolumn{2}{c}{Cityscapes \cite{cordts-cvpr2016-cityscapes}}\\
			\cmidrule[0.5pt]{4-6} \cmidrule[0.5pt]{8-9}
			&  & &(M)$\downarrow$ & \scriptsize{FLOPs(G)$\downarrow$} & FPS$\uparrow$ && \scriptsize{FLOPs(G)$\downarrow$} & FPS$\uparrow$ \\		
			\midrule[0.5pt]
			CutMix\cite{french-bmvc2020-cutmix}    & \scriptsize{BMVC'20}  & DLv2 & 63.86 & 272.33 & 28.47 && 612.74 & 17.30 \\
			\scriptsize{PseudoSeg\cite{zou-iclr2020-pseudoseg}} & ICLR'21  & DLv2 & 63.86 & 272.33 & 28.47 && 612.74 & 17.30 \\
			SDA\cite{yuan-iccv2021-simplebaseline} & ICCV'21  & DLv3+ & 62.62 & 296.06 & 26.20 && 647.40 & 16.23 \\
			ST++\cite{yang-cvpr2022-st++}          & \scriptsize{CVPR'22}  & PSPNet & 65.71 & 262.78 & 28.57 && 591.19 & 17.12 \\
			U$^{2}$PL\cite{wang-cvpr2022-u2pl}     & \scriptsize{CVPR'22}  & DLv3+ & 62.62 & 296.06 & 26.20  && 647.40 & 16.23 \\
			TSS\cite{ke-tip2022-three}             & TIP'22   & DLv2 & 63.86 & 272.33 & 28.47 && 612.74 & 17.30 \\
			CPCL\cite{fan-tip2023-cpcl}            & TIP'23 & DLv3+ & 62.62 & 296.06 & 26.20 && 647.40 & 16.23 \\	
			CCVC\cite{wang-cvpr2023-ccvc}          & \scriptsize{CVPR'23}  & DLv3+ & 62.62 & 296.06 & 26.20 && 647.40 & 16.23 \\
			\midrule[0.5pt]
			TriKD        & 11M  & DF & \textbf{18.28} & \textbf{38.74} & \textbf{78.81} && \textbf{85.48} & \textbf{49.57} \\
			TriKD$^\ast$ & 21M  & DF & \underline{29.02} & \underline{48.86} & \underline{62.74} && \underline{116.88} & \underline{23.47} \\
			\toprule[0.75pt]
		\end{tabular}
	}
\end{table}

\subsection{Quantitative Results}
In semi-supervised setting, we randomly subsample $1/2$, $1/4$, $1/8$, and $1/16$ (the ratio of labeled samples and below the ratio is the number) of images from the training set to construct the pixel-level labeled data, while the rest are used as unlabeled data. The records of the compared methods are taken from the original papers and `-' denotes it is unavailable. The best records are highlighted in boldface and the second-best ones are underlined. We show the results of Pascal VOC2012 \cite{everingham-ijcv2015-pascal} (``VOC'') in Table~\ref{tbl:result_voc} and its augmentation version (``VOCAug'') in Table~\ref{tbl:result_vocaug}, and that of Cityscapes \cite{cordts-cvpr2016-cityscapes} dataset in Table~\ref{tbl:result_cityscapes}.  Also, we provide the inference model size and speed comparison in Table~\ref{tbl:speed_modelPara}.

From these tables, we have the following observations:
\begin{itemize}
	\item Our method with TinyViT-21M (the bottom row) consistently enjoys the most promising performance in terms of mIoU, compared to the state-of-the-art alternatives across all datasets with varying label ratios. For example, TriKD$^\ast$ outperforms the second best CCVC \cite{wang-cvpr2023-ccvc} by 2.83\% on VOCAug with the label ratio of $1/4$ in Table~\ref{tbl:result_vocaug}. This demonstrates the advantages of the proposed triple-view knowledge distillation framework.
	
	\item Our approach has the best precision-speed trade off among all methods. While the performance of TriKD achieves or surpasses the best candidate, it requires much less model parameters and computations. From Table~\ref{tbl:result_voc} and Table~\ref{tbl:speed_modelPara}, TriKD has comparable or better segmentation performance with only one-third (18.28M) or half (29.02M) of the model parameters of the most competitive alternative CCVC \cite{wang-cvpr2023-ccvc} (62.62M). Meanwhile, the required computations are one-sixth (48.86G) or one-seventh (38.74G) of that of CCVC (296.06G) on VOC \cite{everingham-ijcv2015-pascal}, and one-sixth (116.88G) or one-eighth (85.48G) of that of CCVC (647.40G) on Cityscapes \cite{cordts-cvpr2016-cityscapes}, in terms of FLOPs. Moreover, the inference speed of our method achieves 62$\sim$78 fps on VOC and 23$\sim$49 fps on Cityscapes, which are 2$\sim$3 times faster than that of the best candidate. This verifies that our method strikes a good balance between precision and speed, which allows it to be deployed in highly-demanding environment.
	
	\item With the help of plentiful unlabeled data, the segmentation performances are boosted a lot on all the datasets. The largest gains are obtained on VOC \cite{everingham-ijcv2015-pascal} in Table~\ref{tbl:result_voc}, \ie, from 44.32\% to 67\% with TinyViT-11M and from 55.22\% to 70.29\% with TinyViT-21M at the ratio of $1/16$. This validates the fact that a good many unlabeled images may provide complementary spatial knowledge to the model training besides the pixel-level semantics by labeled images. In addition, the segmentation performance is continuously improved with the increasing ratio of labeled samples from $1/16$ to $1/2$.
	
\end{itemize}

\subsection{Ablation Studies}
We examine the components of our TriKD framework, including triple-view encoder, decoder, Knowledge Distillation (KD) loss, and hyper-parameters $\{\lambda_1, \lambda_2, \lambda\}$. The backbone of student network adopts TinyViT-11M with the label ratio of $1/2$, and the remaining parameters are the same as training.

\begin{table}[!t]
	\centering
	\caption{Ablations on triple-view encoder.}
	\label{tbl:ablation_encoder}
	\setlength{\tabcolsep}{1.2mm}{  
    \begin{tabular}{l l c c c c}
	\toprule[0.75pt]
	
	Teacher  & Student & VOCAug & Cityscapes & Params(M) & FLOPs(G) \\		
	\midrule[0.5pt]
	ConvNet  & ConvNet & \underline{79.26} & \underline{75.05} & 57.17 & 91.01 \\
	ViT      & ViT     & 79.21 & 74.51 & 100.74 & 187.12 \\
	hybrid   & hybrid  & 78.08 & 73.02 & 18.28 & 38.74 \\
	ConvNet  & hybrid  & 79.22 & 74.33 & 18.28 & 38.74 \\
	ViT      & hybrid  & 78.87 & 73.92 & 18.28 & 38.74 \\
	\scriptsize{ConvNet+ViT}  & hybrid  & \textbf{80.03} & \textbf{75.98} & 18.28 & 38.74 \\
	\toprule[0.75pt]
    \end{tabular}
}
\end{table}

\textbf{Triple-view encoder}. We investigate the performance when adopting different backbones for teacher network and student network. The results are shown in Table~\ref{tbl:ablation_encoder}, where ``hybrid'' denotes the hybrid ConvNet-ViT, ``Params'' and ``FLOPs'' are computed for the inference phase which only uses student network. From the table, we observe that when both teacher and student networks employ the same backbone ConvNet ($row~1$), it achieves the second best results at the cost of 3.13 times more parameters and 2.35 times larger FLOPs, which are much more worse (\ie, 5.51 times parameters and 4.83 times FLOPs) for adopting ViT ($row~2$) as the backbone. When using hybrid ConvNet-ViT as student network, ConvNet performs the best when only using one teacher network ($row~4$), which indicates that convolutional networks distill the knowledge in a better way than ViT. Most importantly, the performance tops among all settings when using both ConvNet and ViT as teacher networks ($bottom~row$), which validates the superiority of transferring both local and global knowledge to a lightweight student network.

\begin{table}[!t]
	\centering
	\caption{Ablations on knowledge distillation loss. (mIoU \%)}
	\label{tbl:ablation_kdloss}
	\begin{tabular}{c c l  l}
		\toprule[0.75pt]
		$\mathcal{L}_{spa}$ & $\mathcal{L}_{att}$ & VOCAug & Cityscapes \\		
		\midrule[0.5pt]
		           &            & 78.32           & 74.46 \\
		\checkmark &            & 79.23 (+0.91)   & 75.28 (+0.82)\\
		           & \checkmark & 79.18 (+0.86)   & 75.03 (+0.57)\\
		\checkmark & \checkmark & 80.03 (+1.71)   & 75.98 (+1.52) \\
		\toprule[0.75pt]
	\end{tabular}
\end{table}

\textbf{KD loss}. We examine the effects of two knowledge distillation losses, \ie, the spatial loss $\mathcal{L}_{spa}$ and the attention loss $\mathcal{L}_{att}$, and show the results in Table~\ref{tbl:ablation_kdloss}. Compared to the baseline without knowledge distillation, the spatial loss boosts the performance by 0.91\% and 0.82\% in terms of mIoU on VOCAug \cite{everingham-ijcv2015-pascal} and Cityscapes \cite{cordts-cvpr2016-cityscapes}, respectively. The attention loss brings about less gains than that of VOCAug, which suggests that the local spatial structure plays a more important role than the global one for semantic segmentation. When considering both the locality property and the global context by introducing the two losses, the performance is upgraded by 1.71\% and 1.52\% on VOCAug \cite{everingham-ijcv2015-pascal} and Cityscapes \cite{cordts-cvpr2016-cityscapes}, respectively, which verifies the advantage of our KD loss.

\begin{table}[!t]
	\centering
	\caption{Ablations on decoder. (mIoU \%)}
	\label{tbl:ablation_decoder}
	\begin{tabular}{l c c c c}
		\toprule[0.75pt]
		Decoder  & VOCAug           & Cityscapes     & Params(M) & FLOPs(G) \\		
		\midrule[0.5pt]
		Baseline & 79.92            & 75.66          & 39.72     & 221.05 \\
		Ours     & \textbf{80.03}   & \textbf{75.98} & 18.28     & 38.74 \\
		\toprule[0.75pt]
	\end{tabular}
\end{table}

\begin{table}[!t]
	\centering
	\caption{Ablations on hyper-parameter. (mIoU \%)}
	\label{tbl:ablation_hyperpara}
	\setlength{\tabcolsep}{0.6mm}{  
		\begin{tabular}{c c c c c c c c c c c}
			\toprule[0.75pt]
			${\lambda }_{1}$ & VOCAug & Cityscapes && ${\lambda }_{2}$ & VOCAug & Cityscapes && $\lambda$ & VOCAug & Cityscapes\\		
			\midrule[0.5pt]
			0.0 & 79.18 & 75.03 && 0.0 & 79.23 &  75.28 && 0.0 & 76.80 & 71.55 \\
			0.1 & 78.90 & 75.40 && 0.1 & 79.21 & 75.18 && 0.1 & \textbf{80.03} & \textbf{75.98}\\
			0.3 & \underline{79.31} & \underline{75.51} && 0.3 & \underline{79.50} & \underline{75.22} && 0.3 & \underline{79.64}	& \underline{74.69}\\
			0.5 & \textbf{80.03} & \textbf{75.98} && 0.5 & \textbf{80.03} & \textbf{75.98} && 0.5 & 79.42	& 74.36\\
			0.7 & 79.07	& 75.34 && 0.7 & 79.01 & 75.08 && 0.7 & 79.23 & 73.77 \\		
			0.9 & 79.13	& 75.16 && 0.9 & 78.91 & 74.95 && 0.9 & 78.29 & 73.04 \\
			1.0 & 79.02 & 75.11 && 1.0 & 78.32 & 74.67 && 1.0 & 78.03 & 72.98  \\
			\toprule[0.75pt]
		\end{tabular}
	}
\end{table}

\textbf{Decoder}. To show the superiority of our dual-frequency decoder, we make the record of the baseline ($row~1$) that using UPerNet \cite{xiao-eccv2018-upernet} as decoder in Table~\ref{tbl:ablation_decoder}. From the table, we see that our decoder slightly outperforms the baseline at much less cost, only 46.0\% parameters and 17.5\% FLOPs of the baseline. This validates the merits of the dual-frequency design.

\textbf{Loss hyper-parameter}. To examine the contributions of the spatial loss, the attention loss, and the cross pseudo supervision loss, we respectively vary the hyper-parameters $\lambda_1$, $\lambda_2$, and $\lambda$ from 0 to 1 at an interval of 0.1 in Table~\ref{tbl:ablation_hyperpara} (when one hyper-parameter changes, the rest keep still). From the results, it can be found that the performance achieves the best when $\lambda_1$ and $\lambda_2$ are both 0.5, \ie, the spatial loss and the attention loss contributes equally to the model; simultaneously, it obtains the most promising result when $\lambda$ is 0.1 for the cross pseudo supervision loss, which indicates that the weak supervision should not be over-emphasized.

\subsection{Qualitative Results}
\begin{figure}[!t]
	\centering
	\includegraphics[width=0.5\textwidth]{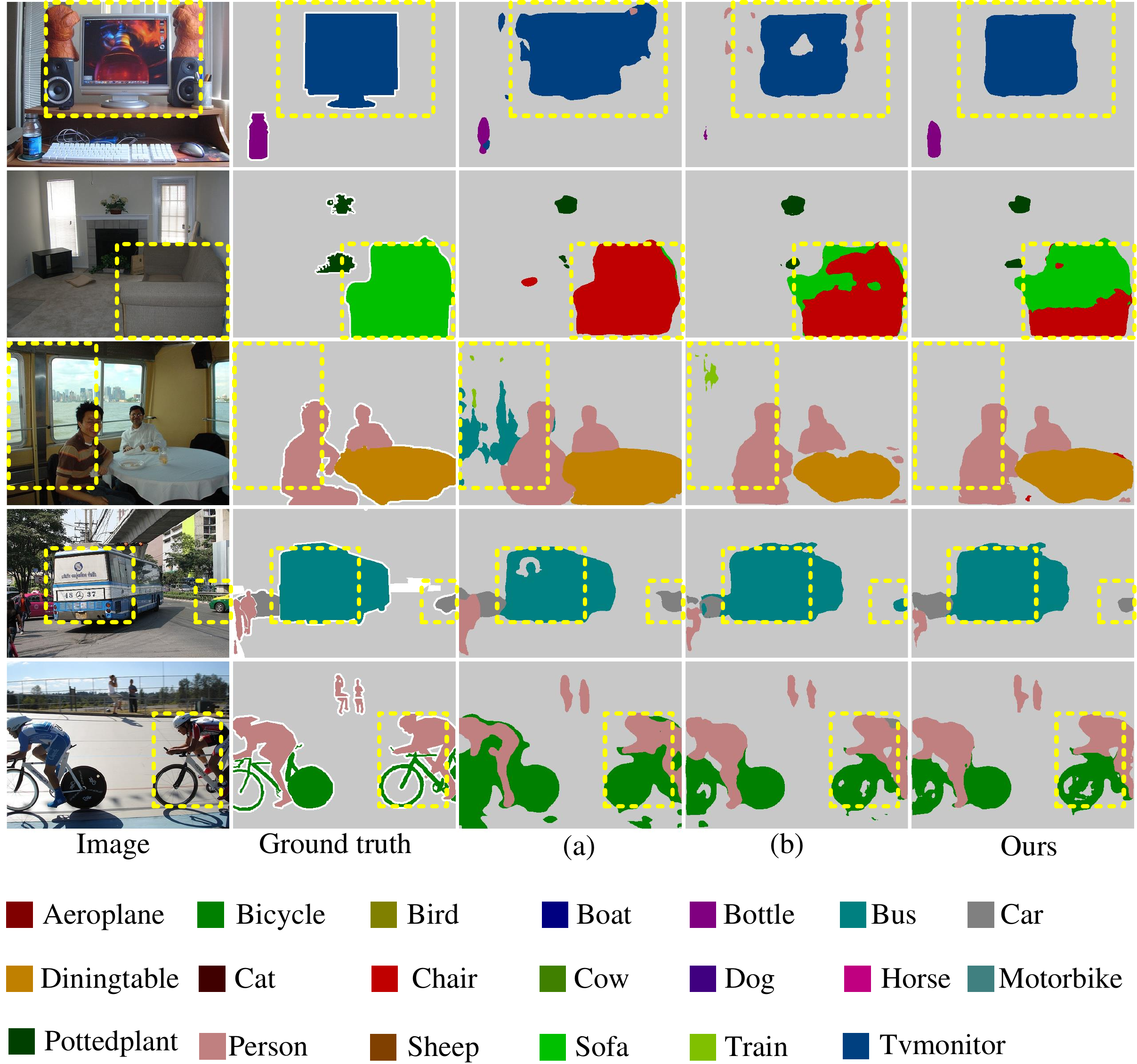}
	\caption{Visualization on VOCAug \cite{everingham-ijcv2015-pascal}. (a) CPS; (b) TriKD$_s^\ast$; (c) TriKD$^\ast$.}
	\label{fig:vocaug_visual}
\end{figure}

\begin{figure}[!t]
	\centering
	\includegraphics[width=0.5\textwidth]{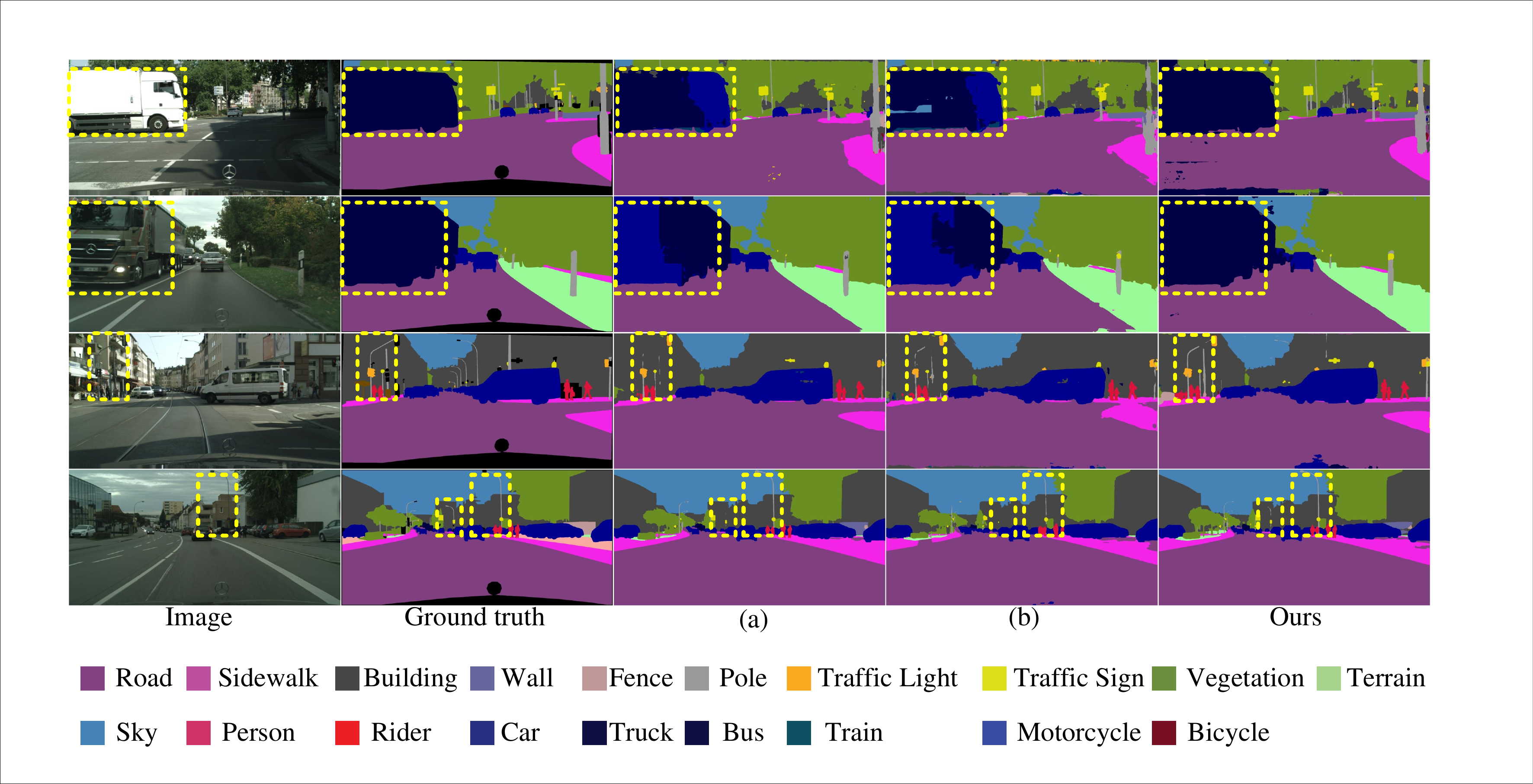}
	\caption{Visualization on Cityscapes \cite{cordts-cvpr2016-cityscapes}. (a) CPS; (b) TriKD$_s^\ast$; (c) TriKD$^\ast$.}
	\label{fig:cityscapes_visual}
\end{figure}

We randomly chose some images from VOCAug \cite{everingham-ijcv2015-pascal} dataset and Cityscapes \cite{cordts-cvpr2016-cityscapes} dataset, and visualize their segmentation results by using different colors to mark the semantic categories in Fig.~\ref{fig:vocaug_visual} and Fig.~\ref{fig:cityscapes_visual}, respectively. For better view, we use the dashed rectangle to highlight the difference zone. As depicted in the two figures, we see that our TriKD method enjoys the most satisfying performance on the two benchmarks compared to its supervised version and CPS \cite{chen-cvpr2021-cps}. In particular, it can well discriminate the tiny objects such as the vehicle wheel ($bottom~row$ in Fig.~\ref{fig:vocaug_visual}) and lamp pole ($row~3\&4$ in Fig.~\ref{fig:cityscapes_visual}) when there exists confusion or crack by the alternatives. The reason may be that the local spatial patterns are preserved by the student network via the knowledge distillation scheme when the global pixel-level context is considered with the help of rich unlabeled samples.

\section{Conclusion}
\label{conclusion}
This paper studies the semi-supervised semantic segmentation task from the perspective of knowledge distillation. To make the student network capture both the local and global context from the teacher network, we design the triple-view encoder by adopting both pure ConvNet and pure ViT as the backbones of teacher networks. Specifically, the introduced spatial loss and attention loss guarantee the low-level local spatial relations and the high-level global semantics to be distilled to the student network. Meanwhile, we modeling the feature channel importance by the channel-wise attention mechanism in the frequency domain. During the inference, we adopt the lightweight student network and dual-frequency decoder for segmentation, which allows our approach to achieve the most promising performances on two benchmarks.

\bibliographystyle{IEEEtran}


\ifCLASSOPTIONcaptionsoff
  \newpage
\fi

\end{document}